\documentclass[10pt,twocolumn,letterpaper]{article}

\usepackage{wacv}
\usepackage{times}
\usepackage{epsfig}
\usepackage{graphicx}
\usepackage{amsmath}
\usepackage{amssymb}

\usepackage{tabularx}
\usepackage{ragged2e}
\usepackage{multirow}
\usepackage{booktabs}

\def\etal{{\em et al.~}}

\newcolumntype{C}{>{\Centering\arraybackslash}X}
\newcolumntype{t}{>{\Centering\hsize=.6\hsize}X}
\newcolumntype{s}{>{\Centering\hsize=.5\hsize}X}
\newcolumntype{j}{>{\Centering\hsize=.4\hsize}X}
\newcolumntype{k}{>{\Centering\hsize=.3\hsize}X}
\newcolumntype{y}{>{\Centering\hsize=.2\hsize}X}
\newcolumntype{i}{>{\Centering\hsize=.09\hsize}X}
\makeatother
\usepackage[normalem]{ulem}
\usepackage{xcolor}

\def\wacvPaperID{576} 

\wacvfinalcopy 
\ifwacvfinal
\usepackage[breaklinks=true,bookmarks=false]{hyperref}
\hypersetup{
    colorlinks,
    linkcolor={black},
    citecolor={black},
    urlcolor={black}
}
\else
\usepackage[pagebackref=true,breaklinks=true,colorlinks,bookmarks=false]{hyperref}
\fi

\ifwacvfinal
\else
\fi

\makeatletter
\def\blfootnote{\xdef\@thefnmark{}\@footnotetext}
\makeatother

\begin{document}

\title{Lip-reading with Densely Connected Temporal Convolutional Networks}

\author{
Pingchuan Ma\textsuperscript{1,*} \hspace{0.2cm}
Yujiang Wang\textsuperscript{1,*,$\dagger$} \hspace{0.2cm} 
Jie Shen\textsuperscript{1,2}  \hspace{0.2cm}
Stavros Petridis\textsuperscript{1,3} \hspace{0.2cm}
Maja Pantic\textsuperscript{1,2} \\
\textsuperscript{1}Imperial College London  \hspace{0.33cm} \textsuperscript{2}Facebook London \hspace{0.33cm} \textsuperscript{3}Samsung AI Center Cambridge\\
{\tt\small \{pingchuan.ma16,yujiang.wang14,jie.shen07,stavros.petridis04,m.pantic\}@imperial.ac.uk}
}

\maketitle

\begin{abstract}
\blfootnote{\noindent* Equal Contribution.} 
\blfootnote{$^{\dagger}$ Corresponding author.}
\blfootnote{An improved code implementation is available at: \url{https://github.com/mpc001/Lipreading_using_Temporal_Convolutional_Networks}
}
In this work, we present the Densely Connected Temporal Convolutional Network (DC-TCN) for lip-reading of isolated words. Although Temporal Convolutional Networks (TCN) have recently demonstrated great potential in many vision tasks, its receptive fields are not dense enough to model the complex temporal dynamics in lip-reading scenarios. To address this problem, we introduce dense connections into the network to capture more robust temporal features. Moreover, our approach utilises the Squeeze-and-Excitation block, a light-weight attention mechanism, to further enhance the model's classification power. Without bells and whistles, our DC-TCN method has achieved 88.36\,\% accuracy on the Lip Reading in the Wild (LRW) dataset and 43.65\,\% on the LRW-1000 dataset, which has surpassed all the baseline methods and is the new state-of-the-art on both datasets.

\end{abstract}

\section{Introduction}
Visual Speech Recognition, also known as lip-reading, consists of the task of recognising a speaker's speech content from visual information alone, typically the movement of the lips. Lip-reading can be extremely useful under scenarios where the audio data is unavailable, and it has a broad range of applications such as in silent speech control system \cite{sun2018lip}, for speech recognition with simultaneous multi-speakers and to aid people with hearing impairment. In addition, lip-reading can also be combined with an acoustic recogniser to improve its accuracy.     

Despite of many recent advances, lip-reading is still a challenging task. Traditional methods usually follow a two-step approach. The first stage is to apply a feature extractor such as Discrete Cosine Transform (DCT) \cite{hong2006pca,potamianos2003recent,potamianos2001cascade} to the mouth region of interest (RoI), and then feed the extracted features into a sequential model (usually a Hidden Markov Model or HMM in short) \cite{estellers2011dynamic,shao2008stream,dupont2000audio} to capture the temporal dynamics. Readers are referred to \cite{zhou2014review} for more details about these older methods. 

The rise of deep learning has led to significant improvement in the performance of lip-reading methods. Similar to traditional approaches, the deep-learning-based methods usually consist of a feature extractor (front-end) and a sequential model (back-end). Autoencoder models were applied as the front-end in the works of \cite{gehring2013extracting,noda2015audio,petridis2016deep} to extract deep bottleneck features (DBF) which are more discriminative than 
DCT features. Recently, the 3D-CNN (typically a 3D convolutional layer followed by a deep 2D Convolutional Network) has gradually become a popular front-end choice \cite{stafylakis2017combining, martinez2020lipreading, petridis2018end, weng2019learning}. As for the back-end models, Long-Short Term Memory (LSTM) networks were applied in \cite{petridis2016deep,stafylakis2017combining,petridis2017end,petridis2017end} to capture both global and local temporal information. Other widely-used back-end models includes the attention mechanisms \cite{chung2017lip,petridis2018audio}, self-attention modules \cite{afouras2018deep}, and Temporal Convolutional Networks (TCN) \cite{bai2018empirical, martinez2020lipreading}.

Unlike Recurrent Neural Networks (RNN) such as LSTMs or Gated Recurrent Units (GRUs) \cite{cho2014learning} with recurrent structures and gated mechanisms, Temporal Convolutional Networks (TCN) adopt fully convolutional architectures and have the advantage of faster converging speed with longer temporal memory. The authors of \cite{bai2018empirical} described a simple yet effective TCN architecture which outperformed baseline RNN methods, suggesting that TCN can be a reasonable alternative for RNNs on sequence modelling problems. Following this work, it was further demonstrated in \cite{martinez2020lipreading} that a multi-scale TCN could achieve better performance than RNNs on lip-reading of isolated words, which is also the state-of-the-art model so far. Such multi-scale TCN stacks the outputs from convolutions with multiple kernel sizes to gain a more robust temporal features, which has already been shown to be effective in other computer vision tasks utilising multi-scale information such as the semantic segmentation \cite{chen2017deeplab, zhao2017pyramid,chen2017rethinking}. The TCN architectures in both works \cite{bai2018empirical,martinez2020lipreading} have adopted dilated convolutions \cite{yu2015multi} to enlarge the receptive fields of models. Under the scenarios of lip-reading, a video sequence usually contains various subtle syllables that are essential to distinguish the word or sentence, and thus the model's abilities to compactly cover those syllables are necessary and important. However, TCN architectures in \cite{bai2018empirical,martinez2020lipreading} have utilised a sparse connection and thus may not observe the temporal features thoroughly and densely.


Inspired by recent success of Densely Connected Networks \cite{huang2017densely,yang2018denseaspp,guo2019dense}, we introduce dense connections into the TCN structures and propose the Densely Connected TCN (DC-TCN) for word-level lip-reading. DC-TCNs are able to cover the temporal scales in a denser fashion and thus are more sensitive to words that may be challenging to previous TCN architectures \cite{bai2018empirical,martinez2020lipreading}. Specifically, we explore two approaches of adding dense connections in the paper. One is a fully dense (FD) TCN model, where the input of each Temporal Convolutional (TC) layers is the concatenations of feature maps from all preceding TC layers. Another DC-TCN variant employs a partially dense (PD) structures. We further utilise the Squeeze-and-Excitation (SE) attention machanism\cite{hu2018squeeze} in both DC-TCN variants, which further enhances their classification power.

To validate the effectiveness of the proposed DC-TCN models, we have conducted experiments on the Lip Reading in the Wild (LRW) \cite{chung2016lip} dataset and LRW-1000 dataset \cite{yang2019lrw}, which are the largest publicly available benchmark datasets for unconstrained lip-reading in English and in Mandarin, respectively. Our final model achieves 88.36\,\% accuracy on LRW, surpassing the current state-of-the-art method \cite{martinez2020lipreading} (85.3\,\%) by around
3.1\,\%. On LRW-1000, our method gains 43.65\,\% accuracy and also out-perform all baselines, demonstrating the generality and strength of the proposed DC-TCN. 

In general, this paper presents the following contributions :

1. We propose a Densely Connected Temporal Convolutional Network (DC-TCN) for lip-reading of isolated words, which can provide denser and more robust temporal features.

2. Two DC-TCN variants with Squeeze-and-excitation blocks \cite{hu2018squeeze}, namely the fully dense (FD) and partially dense (PD) architectures, are introduced and evaluated in this paper.

3. Our method has achieved 88.36\,\% top-1 accuracy on LRW dataset and 43.65\,\% and on LRW-1000, which have surpassed all baseline methods and set a new record on these two datasets.
\section{Related Works}
\subsection{Lip-reading}
Early deep learning methods for lip-reading of isolated words were mainly evaluated on small-scale datasets recorded in constrained environments such as OuluVS2 \cite{anina2015ouluvs2} and CUAVE \cite{patterson2002moving}. The authors of \cite{petridis2016deep} proposed to use the combination of deep bottleneck features (DBF) and DCT features to train a LSTM classifier, which is not end-to-end trainable. An end-to-end trainable model was first demonstrated in \cite{wand2016lipreading} using Histogram of Oriented Gradient (HOG) features with LSTMs, and later Petridis \etal \cite{petridis2017end} trained an end-to-end model with a Bidirectional LSTM back-end where the first and second derivatives of temporal features are also computed, achieving much better results than \cite{wand2016lipreading}. The lip-reading accuracy on those small-scale datasets are further improvded by the introduction of multi-view visual information \cite{petridis2017end} and audio-visual fusion \cite{petridis2017end2, petridis2020end}. 

Lip Reading in the Wild (LRW) dataset \cite{chung2016lip} is the first and the largest publicly available dataset for unconstrained lip-reading with a 500-word English vocabulary. It has encouraged the emergence of numerous deep learning models with more and more powerful word-recognising abilities. The WLAS sequence-to-sequence model \cite{chung2017lip} consists of a VGG network \cite{simonyan2014very} and a LSTM with dual attention systems on visual and audio stream, respectively. LipNet \cite{assael2016lipnet} is the first approach to employ a 3D-CNN to extract spatial-temporal features that are classified by Bidirectional Gated Recurrent Units (BGRUs). A 2D Residual Network (ResNet) \cite{he2016deep} on top of a 3D Convolutional layer is used as the front-end in \cite{stafylakis2017combining} (with an LSTM as the back-end). Two 3D ResNets are organised in a two-stream fashion (one stream for image and another for optical flow) in the work of \cite{weng2019learning}, learning more robust spatial-temporal features at the cost of larger network size. Cheng \etal \cite{DBLP:conf/icassp/00010TPB0P20} propose a technique of pose augmentation to enhance the performance of lip-reading in extreme poses. Zhang \etal \cite{zhang2020can} propose to incorporate other facial parts in additional to the mouth region to solve lip-reading of isolated words, and mutual information constrains are added in \cite{zhao2020mutual} to produce more discriminative features. The current state-of-the-art performance on LRW is achieved by \cite{martinez2020lipreading}, which has replaced RNN back-ends with a multi-scale Temporal Convolutional Networks (TCN). The same model achieves the state-of-the-art performance on LRW-1000 \cite{yang2019lrw} dataset which is currently the largest lip-reading dataset for Mandarin. 

\subsection{Temporal convolutional networks}
Although RNN networks such as LTSMs or GRUs had been commonly used in lip-reading methods to model the temporal dependencies, alternative light-weight, faster-converging CNN models have started to gain attention in recent works. Such efforts can be traced back to the Time-Delay Neural Networks (TDNN) \cite{waibel1989phoneme} in 1980s. Consequently, models with Temporal Convolutions were developed, including WaveNet \cite{oord2016wavenet} and Gated ConNets \cite{dauphin2017language}. Lately, Bai \textit{et al.} \cite{bai2018empirical} described a simple and generic Temporal Convolutional Network (TCN) architecture that out-performed baseline RNN models in various sequence modelling problems. 
Although the TCN introduced in \cite{bai2018empirical} is a causal one in which no future features beyond the current time step can be seen in order to prevent the leakage of future information, the model can also be modified into a non-causal variant without such constrains. The work of \cite{martinez2020lipreading} has adopted a non-casual TCN design, where the linear TC block architecture was replaced with a multi-scale one. To the best of our knowledge, this work \cite{martinez2020lipreading} has achieved the current state-of-the-art performance on the LRW dataset. However, the receptive field scales in such TCN architectures may not be able cover the full temporal range under lip-reading scenarios, which can be solved by the employment of dense connections.  

\subsection{Densely connected networks}
Densely connected network has received broad attention since its inception in \cite{huang2017densely}, where a convolutional layer receives inputs from all its preceding layers. Such densely connected structure can effectively solve the vanishing-gradient problem by employing shallower layers and thus benefiting feature propagation. The authors of \cite{yang2018denseaspp} have applied dense connections to dilated convolutions to enlarge the receptive field sizes and to extract denser feature pyramid for semantic segmentation. Recently, a simple dense TCN for Sign Language Translation has been illustrated in \cite{guo2019dense}. Our work is the first to explore the densely connected TCN for for word-level lip-reading, where we present both a fully dense (FD) and a partially dense (PD) block architectures with the addition of the channel-wise attention method described in \cite{hu2018squeeze}.

\subsection{Attention and SE blocks}
Attention mechanism \cite{bahdanau2014neural, luong2015effective, vaswani2017attention, wang2018non} can be used to teach the network to focus on the more informative locations of input features. In lip-reading, attention mechanisms have been mainly developed for sequence models like LSTMs or GRUs. A dual attention mechanism is proposed in \cite{chung2017lip} for the visual and audio input signals of the LSTM models. Petridis et al. \cite{petridis2018audio} have coupled the self-attention \cite{vaswani2017attention} block with a CTC loss to improve the performance of Bidirectional LSTM classifiers. Those attention methods are somehow computational expensive and are inefficient to be integrated into TCN structures. In this paper, we adopt a light-weight attention block, which is the Squeeze-and-Excitation (SE) network \cite{hu2018squeeze}, to introduce the channel-wise attention into the DC-TCN network. 

In particular, denote the input tensor of a SE block as $U \in \mathbb{R}^{C \times H \times W}$ where $C$ is the channel number, its channel-wise descriptor $z \in \mathbb{R}^{C \times 1 \times 1}$ is first be obtained by a global average pooling operation to squeeze the spatial dimension ${H \times W}$, i.e. $z = GlobalPool(U)$ where $GlobalPool$ denotes the global average pooling. After that, an excitation operation is applied to $z$ to obtain the channel-wise dependencies $s \in \mathbb{R}^{C \times 1 \times 1}$, which can be expressed as $s = \sigma (W_u \delta (W_v z))$. Here, $W_v \in \mathbb{R}^{\frac{C}{r} \times C}$ and $W_u \in \mathbb{R}^{C \times  \frac{C}{r}}$ are learnable weights, while $\sigma$ and $\delta$ stands for the sigmoid activation and ReLU functions and $r$ represents the reduction ratio. The final output of the SE block is simply the channel-wise broadcasting multiplication of $s$ and $U$. The readers are referred to \cite{hu2018squeeze} for more details. 

\section{Methodology}
\subsection{Overview}
\begin{figure}[t!]
  \centering
  \includegraphics[width=0.45\linewidth]{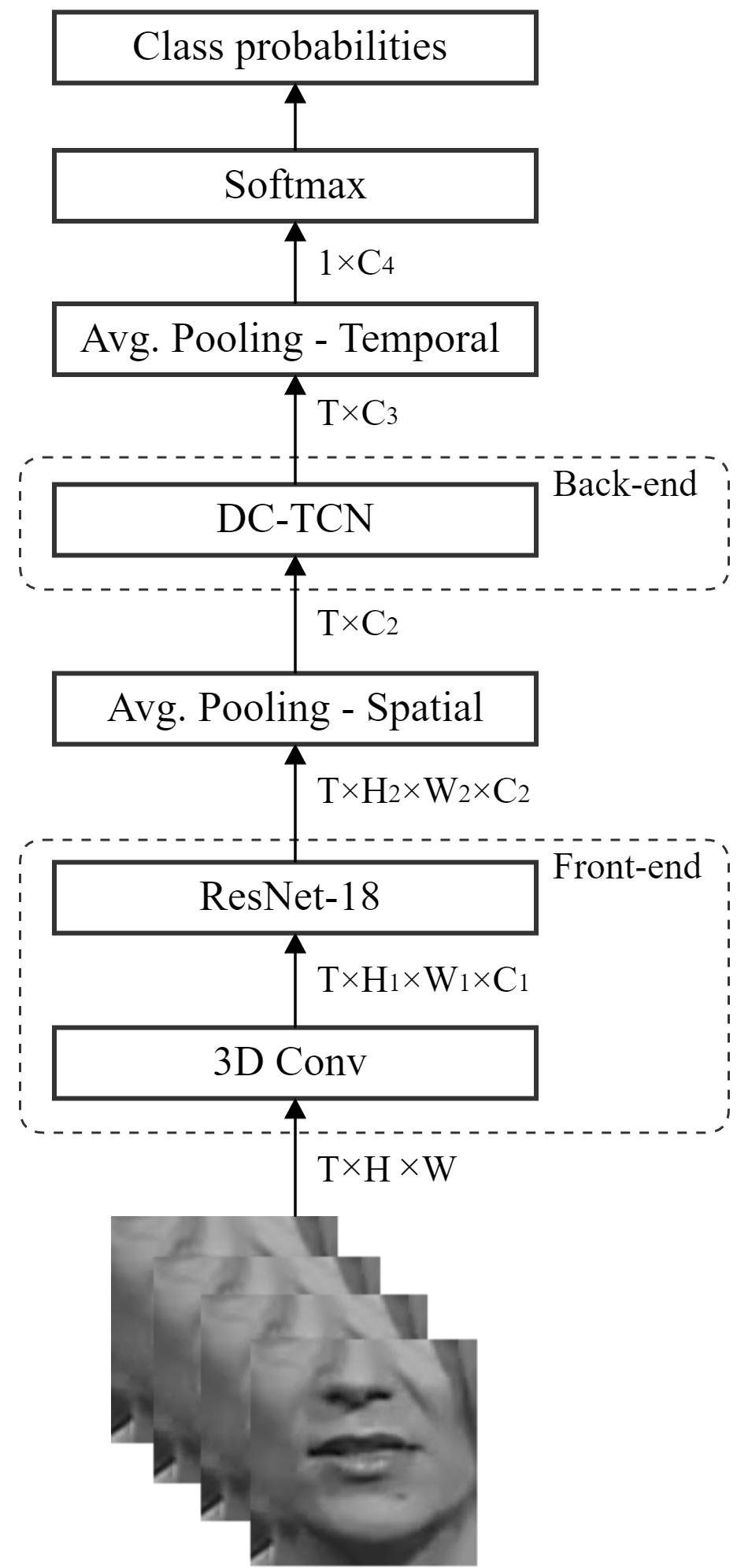}
  \caption{The general framework of our method. We utilise a 3D convolutional Layer plus a 2D ResNet-18 to extract features from the input sequence, while the proposed Densely Connected TCN (DC-TCN) models the temporal dependencies. $C_1$, $C_2$, $C_3$ denotes different channel numbers, while $C_4$ refers to the total word classes to be predicted. The batch size dimension is ignored for simplicity. }
  \label{framework}
  \vspace{-2mm}
\end{figure}
Fig. \ref{framework} depicts the general framework of our method. The input is the cropped gray-scale mouth RoIs with the shape of $T\times H \times W$, where $T$ stands for the temporal dimension and $H$, $W$ represent the height and width of the mouth RoIs, respectively. Note that we have ignored the batch size for simplicity. Following \cite{stafylakis2017combining, martinez2020lipreading}, we first utilise a 3D convolutional layer to obtain the spatial-temporal features with shape $T\times H_1\times W_1\times C_1$, where $C_1$ is the feature channel number. On top of this layer, a 2D ResNet-18 \cite{he2016deep} is applied to produce features with shape $T\times H_2\times W_2\times C_2$. The next layer applies the average pooling to summarise the spatial knowledge and to reduce the dimensionality to $T\times C_2$. After this pooling operation, the proposed Densely Connected TCN (DC-TCN) is employed to model the temporal dynamics. The output tensor ($T\times C_3$) is passed through another average pooling layer to summarise temporal information into $C_4$ channels, while $C_4$ represents the classes to be predicted. The word class probabilities are predicted by the succeeding softmax layer. The whole model is end-to-end trainable.

\subsection{Densely Connected TCN}
\label{DC-TCN structures}

\begin{figure}[t!]
  \centering
  \includegraphics[width=0.6\linewidth]{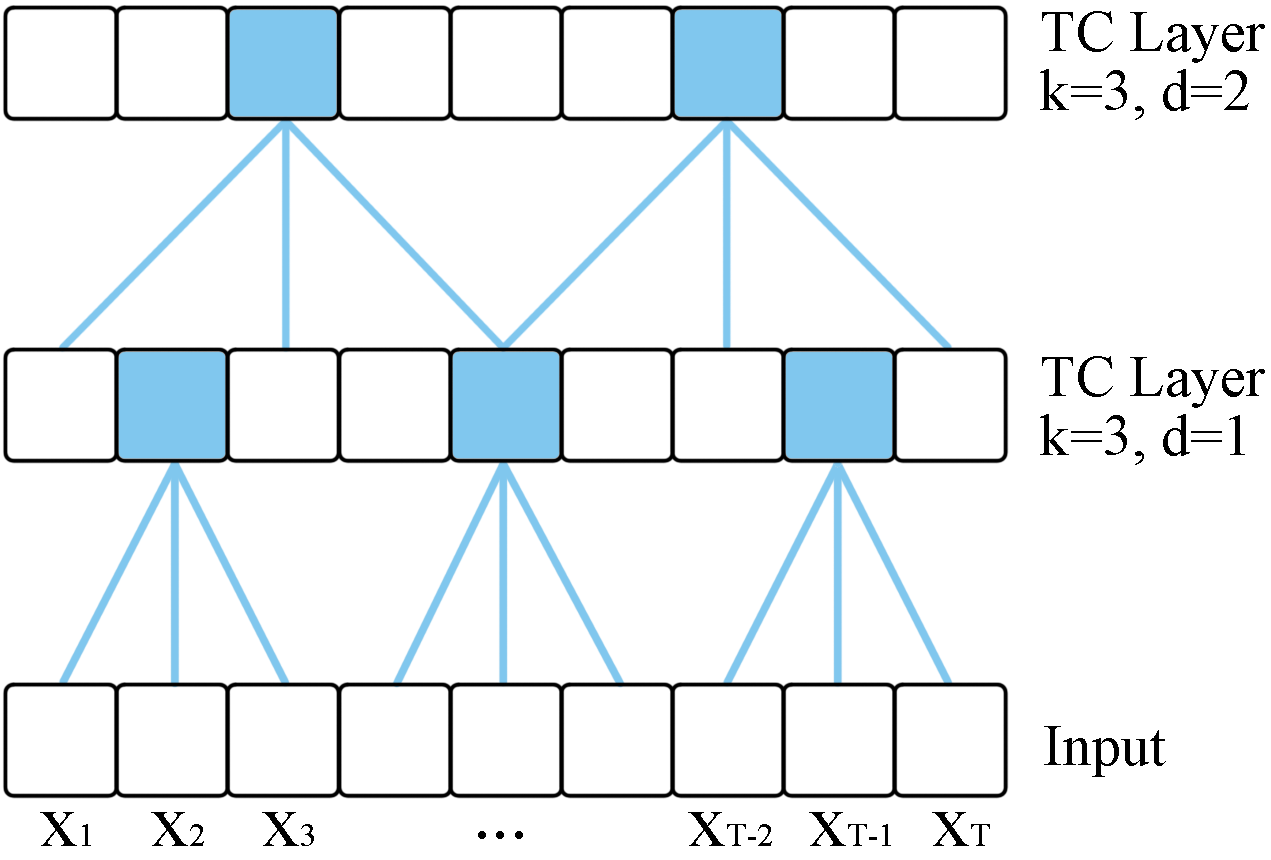}
  \caption{An illustration of the non-causal temporal convolution layers where $k$ is the filter size and $d$ is the dilation rate. The receptive fields for the filled neurons are shown.}
  \label{tc_example}
  \vspace{-2mm}
\end{figure}

\begin{figure*}[ht!]
  \centering
  \includegraphics[width=0.76\linewidth]{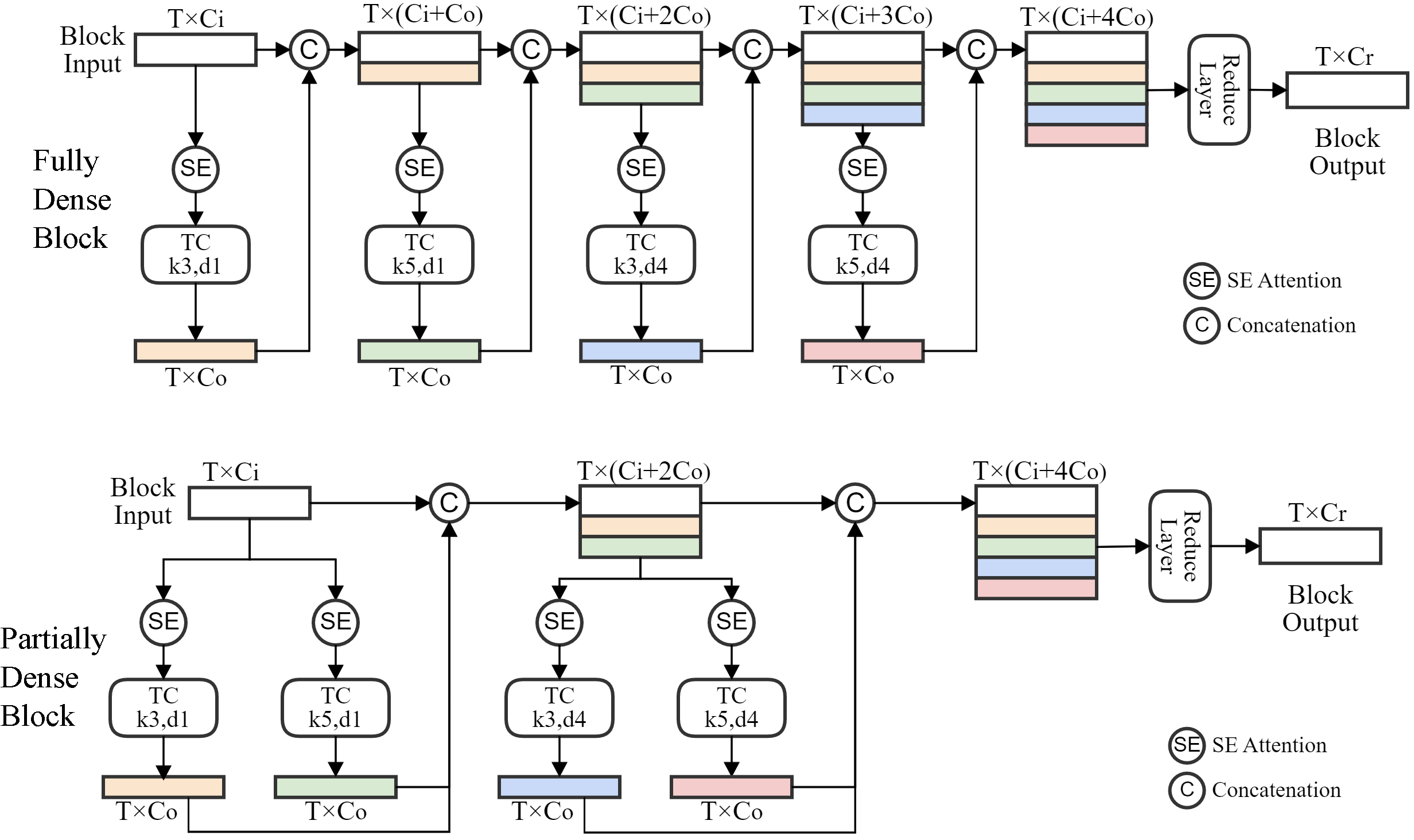}
  \caption{ The architectures of the fully dense block (Up) and the partially dense block (bottom) in DC-TCN. We have selected the block filter sizes set $K = \{3,5\}$ and the dilation rates set $D = \{1,4\}$ for simplicity. In both blocks, Squeeze-and-Excitation (SE) attention is attached after each input tensor. A reduce layer is involved for channel reduction.}
  \label{block_architecture}
\end{figure*}

To introduce the proposed Densely Connected TCN (DC-TCN), we start from a brief explanation of the temporal convolution in \cite{bai2018empirical}. A temporal convolution is essentially a 1-D convolution operating on temporal dimensions, while a dilation \cite{yu2015multi} is usually inserted into the convolutional filter to enlarge the receptive fields. Particularly, for a 1-D feature $\mathbf{x} \in \mathbb{R}^{T}$ where $T$ is the temporal dimensionality, 
define a discrete function $g : \mathbb{Z^+} \mapsto \mathbb{R}$ such that $g(s) = \mathbf{x}_s$ where $s \in [1,T] \cap \mathbb{Z}^+$, let $\Lambda_k=[1,k] \cap \mathbb{Z}^+$ and $f : {\Lambda_k} \mapsto \mathbb{R}$ be a 1D discrete filter of size $k$, a temporal convolution $*_d$ with dilation rate $d$ is described as
\begin{equation}
    \label{eq1}
  \mathbf{y}_p = (\mathbf{g} *_d f)(p) = \sum_{s+dt=p} g(s) f(t)
\end{equation}
where $\mathbf{y} \in \mathbb{R}^T$ is the 1-D output feature and $\mathbf{y}_p$ refers to its $p$-th element. Note that zero padding is used to keep the temporal dimensionality unchanged in $\mathbf{y}$. Note that the temporal convolution described in Eq. \ref{eq1} is non-casual, i.e. the filters can observe features of every time step, similarly to that of \cite{martinez2020lipreading}. Fig. \ref{tc_example} has provided a intuitive example of the non-casual temporal convolution layers. 

Let $TC^l$ be the $l$-th temporal convolution layer, and let $\mathbf{x}^l \in \mathbb{R}^{T \times C_{i}}$ and $\mathbf{y}^l \in \mathbb{R}^{T \times C_{o}}$ be its input and output tensors with $C_{i}$ and $C_{o}$ channels, respectively, i.e. $\mathbf{y}^l=TC^l(\mathbf{x}^l)$. In common TCN architectures, $\mathbf{y}^l$ is directly fed into the $(l+1)$-th temporal convolution layer $TC^{l+1}$ to produce its output $\mathbf{y}^{l+1}$, which is depicted as
\begin{equation}
\label{eq2}
\begin{split}
    & \mathbf{x}^{l+1} = \mathbf{y}^{l} \\
    & \mathbf{y}^{l+1} = TC^{l+1} (\mathbf{x}^{l+1}).
\end{split}
\end{equation}

In DC-TCN, dense connections \cite{huang2017densely} are utilised and the input for the following TC layer ($TC^{l+1}$) is the concatenation between $\mathbf{x}^l$ and $\mathbf{y}^l$, which can be written as
\begin{equation}
    \label{eq3}
    \begin{split}
    & \mathbf{x}^{l+1} = [\mathbf{x}^{l}, \mathbf{y}^{l}] \\
    & \mathbf{y}^{l+1} = TC^{l+1} (\mathbf{x}^{l+1}).
\end{split}
\end{equation}
Note that $\mathbf{x}^{l+1} \in \mathbb{R}^{T \times (C_{i}+C_{o})}$ has additional channels ($C_{o}$) than $\mathbf{x}^{l}$, where $C_{o}$ is defined as the growth rate following \cite{huang2017densely}.

We have embedded the dense connections in Eq. \ref{eq3} to constitute the block of DC-TCN. More formally, we define a DC-TCN block to consist of temporal convolution (TC) layers with arbitrary but unique combinations of filter size $k \in K$ and dilation rate $r \in D$, where $K$ and $D$ stand for the sets of all available filter sizes and dilation rates for this block, respectively. For example, if we define a block to have filter sizes set $K = \{3,5\}$ and dilation rates set $D = \{1,4\}$, there will be four TC layers ($k3d1, k5d1, k3d4, k5d4$) in this block. 

In this paper, we study two approaches of constructing DC-TCN blocks. The first approach applies dense connections for all TC layers, which is denoted as the fully dense (FD) block, as illustrated at the top of Fig. \ref{block_architecture}, where the block filter sizes set $K = \{3,5\}$ and the dilation rates set $D = \{1,4\}$. As shown in the figure, the output tensor of each TC layer is consistently concatenated to the input tensor, increasing the input channels by  $C_0$ (the growth rate) each time. Note that we 
have a Squeeze-and-Excitation (SE) block \cite{hu2018squeeze} after the input tensor of each TC layer to introduce channel-wise attentions for better performance. Since the output of the top TC layer in the block typically has much more channels than the block input (e.g. $C_i+4C_0$ channels in Fig. \ref{block_architecture}), we employ a 1$\times$1 convolutional layer to reduce its channel dimensionality from $C_i+4C_0$ to $C_r$ for efficiency (``Reduce Layer'' in Fig. \ref{block_architecture}). A 1$\times$1 convolutional layer is then applied to convert the block input's channels if $C_i \neq C_r$. In the fully dense architecture, TC layers are stacked in a receptive-field-ascending order.


 
Another DC-TCN block design is depicted at the bottom of Fig. \ref{block_architecture}, which we denote as the partially dense (PD) block. In the PD block, filters with identical dilation rates are employed in a multi-scale fashion, such as the $k3d1$ and $k5d1$ TC layers in Fig. \ref{block_architecture} (bottom), and their outputs are concatenated to the input simultaneously.
PD block is a essentially a hybrid of the multi-scale architectures and densely connected networks, and thus is expected to benefit from both. Just like in FD architectures, SE attention is also attached after every input tensor, while the ways of utilising the reduce layer is the same to that of fully dense blocks.

A DC-TCN block, either fully or partially dense, can be seamlessly stacked with another block to obtain finer features. A fully dense / partially dense DC-TCN model can be formed by stacking $B$ identical FD / PD blocks together, where $B$ denotes the number of blocks. 

There are various important network parameters to be determined for a DC-TCN model, including the filter sizes $K$ and the dilation rates $D$ in each block, the growth rate $C_o$, and the reduce layer channel $C_r$ and the blocks number $B$. The optimal DC-TCN architecture along with the process of determining it can be found in Sec. \ref{net_struct_explore}.

\subsection{Advantages of DC-TCN}

\begin{figure}[t!]
  \centering
  \includegraphics[width=0.85\linewidth]{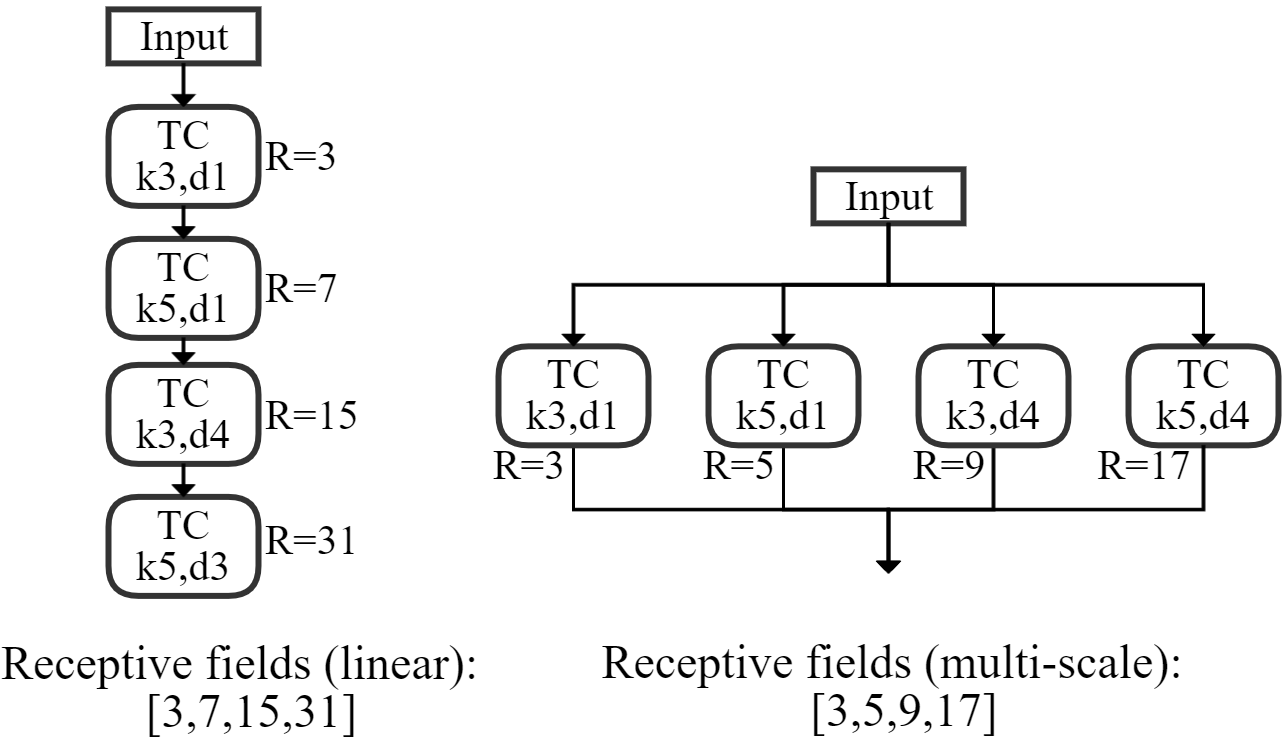}
  \caption{The block receptive field size when combining four TC layers (with a receptive field size of $3,5,9$ and $17$) in a linear (left) or in a multi-scale (right) method.
}
  \label{baseline_rf}
  \vspace{-2mm}
\end{figure}

The receptive field size $R$ for a filter with kernel size $k$ and dilation rate $d$ can be calculated as
\begin{equation}
    \label{eq4}
R = k + (d-1)(k-1).
\end{equation}
Stacking two TC layers with receptive fields $R_1$ and $R_2$ will produce a receptive field size of $(R_1+R_2-1)$. The receptive field sizes for the four TC layers described in Fig. \ref{block_architecture} are $3,5,9$ and $17$, respectively. If they are connected linearly as in \cite{bai2018empirical}, the resulting model can see a temporal range of $(3,7,15,31)$. A multi-scale structure \cite{martinez2020lipreading} will lead to receptive fields of $(3,5,9,17)$. The linearly connected architecture retains a larger maximum receptive size than the multi-scale one, however, it also generate more sparse temporal features. We have illustrated the receptive fields for these two block architectures in Fig. \ref{baseline_rf}.

\begin{figure}[t!]
  \centering
  \includegraphics[width=0.85\linewidth]{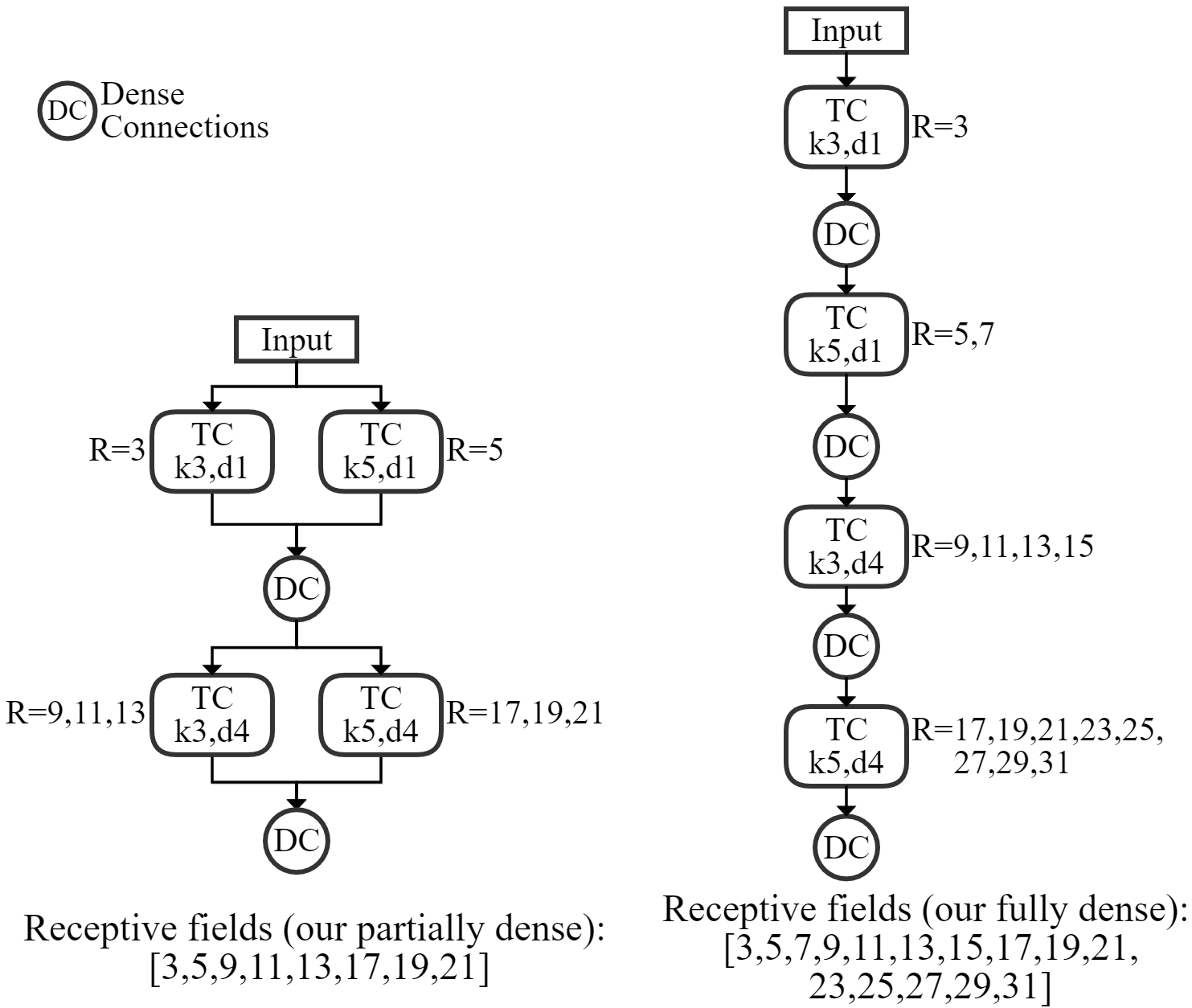}
  \caption{The block receptive field sizes when combining four TC layers (with a receptive field size of $3,5,9$ and $17$) in our partially dense (left) or fully dense (right) block. The dense connection described in Eq. \ref{eq3} is denoted as ``DC''. 
  Compared with the structures in Fig. \ref{baseline_rf}, our DC-TCN can observe denser temporal scales without shrinking the maximum receptive field and thus can produce more robust features.}
  \label{ours_rf}
  \vspace{-2mm}
\end{figure}

Unlike the linearly connected \cite{bai2018empirical} or multi-scale \cite{martinez2020lipreading} TCN, our DC-TCN can extract the temporal features at denser scales and thus increase the features' robustness without reducing the maximum receptive field size. Fig. \ref{ours_rf} depicts the temporal range covered by our partially dense and fully dense blocks, which consist of the four identical TC layers as shown in Fig. \ref{baseline_rf}. Since we have introduced dense connection (``DC'' in the figure) into the structure, a TC layer can see all the preceding layers and therefore the varieties of its receptive sizes are significantly enhanced. As shown in Fig. \ref{ours_rf} (left), our partially dense block can observe a total of eight different ranges, which is double of that in linear or multi-scale architectures (only 4 scales). The fully dense block in Fig. \ref{ours_rf} (right) can produce feature pyramid from 15 different receptive fields with the maximum one to be 31 (larger than multi-scale and equal to linear). Such dense receptive fields can ensure that the information from a wide range of scales can be observed by the model, and thus strengthen the model's expression power. 

\section{Experiments}
\subsection{Datasets}
We have conducted our experiments on the Lip Reading in the Wild (LRW) \cite{chung2016lip} and the LRW-1000 \cite{yang2019lrw} dataset, which are the largest publicly available dataset for lipreading of isolated words in English and in Mandarin, respectively. 

The LRW dataset has a vocabulary of 500 English words. The sequences in LRW are captured from more than 1\,000 speakers in BBC programs, and each sequence has a duration of 1.16 seconds (29 video frames). There are a total of 538\,766 sequences in this dataset, which are split into 488\,766/25\,000/25\,000 for training/validation/testing usages. This is a challenging dataset due to the large number of subjects and variations in head poses and lighting conditions.

The LRW-1000 dataset contains a total of 718\,018 samples for 1\,000 mandarin words, recorded from more than 2\,000 subjects. The average duration for each sequence is 0.3 second, and the total length of all sequences is about 57 hours. The training/validation/testing splits consist of 603\,193/63\,237/51\,588 samples, respectively. This dataset is even more challenging than LRW considering its huge variations in speaker properties, background clutters, scale, etc.

\subsection{Experimental setup}
\textbf{Pre-processing}\quad
We have pre-processed the LRW dataset following the same method as described in \cite{martinez2020lipreading}. We first detect 68 facial landmarks using dlib~\cite{king2009dlib}. Based on the coordinates of these landmarks, the face images are warped to a reference mean face shape. Finally, the mouth RoI of size 96$\times$96 is cropped from each warped face image and is converted into grayscale. For LRW-1000, we simply use the provided pre-cropped mouth RoIs and resize them to 122$\times$122 following \cite{yang2019lrw}.

\textbf{Evaluation metric}\quad
Top-1 accuracy is used to evaluate the model performance, since we are solving word-level lip-reading classification problems. 

\textbf{Training settings} \quad
The whole model in Fig. \ref{framework}, including the proposed DC-TCN, is trained in an end-to-end fashion, where the weights are randomly initialised. We employ identical training settings for both LRW and LRW-1000 datasets except of some slight differences to cope with their different input dimensions. We train 80 epochs with a batch size of 32/16 on LRW/LRW-1000, respectively, and measure the top-1 accuracy using the validation set to determine the best-performing checkpoint weights. We adopt AdamW \cite{loshchilov2017decoupled} as the optimiser, where the initial learning rate is set to 0.0003/0.0015 for LRW and LRW-1000, respectively. A cosine scheduler \cite{loshchilov2016sgdr} is used to steadily decrease the learning rate from the initial value to 0. BatchNorm layers \cite{ioffe2015batch} are embedded to accelerate training convergence, and we use dropouts with dropping probabilities 0.2 for regularisation. The reduction ratio in the SE block is set to 16, and the channel value $C_2$ of DC-TCN's input tensor is set to 512. Besides, we adopt the variable length augmentation as proposed in \cite{martinez2020lipreading} to increase the model's temporal robustness.

\textbf{Explorations of DC-TCN structures}\quad
We evaluate DC-TCN with different structure parameters on LRW dataset to determine the best-performing one. In particular, we first validate the effectiveness of different filter sizes $K$ and dilation rates $D$ in each DC-TCN block while freezing other hyper-parameters such as the growth rate $C_o$ and the reduce layer channels $C_r$. Then we select the most effective $K$ and $D$ values to fine-tune other structural options, including the growth rate $C_o$ and whether to use SE attention. We explore structures for both FD and PD blocks. 

\textbf{Baseline methods}\quad
On the LRW dataset, the performance of the proposed DC-TCN model is compared with the following baselines: 1) the method proposed in the LRW paper \cite{chung2016lip} with a VGG backbone \cite{simonyan2014very}, 2) the WLAS model \cite{chung2017lip}, 3) the work of \cite{stafylakis2017combining} where a ResNet \cite{he2016deep} and a LSTM is used, 4) the End-to-End Audio-Visual network \cite{petridis2018end}, 5) the multi-grained spatial-temporal model in \cite{wang2019multi}, 6) the two-stream 3D CNN in \cite{weng2019learning}, 7) the Global-Local Mutual Information
Maximisation method in \cite{zhao2020mutual}, 8) the face region cutout approach by authors of \cite{zhang2020can}, 9) the multi-modality speech recognition method in \cite{xu2020discriminative}, and 10) the multi-scale TCN proposed by \cite{martinez2020lipreading}. 

LRW-1000 is a relatively new dataset and there are somehow fewer works on it. We have selected the following methods as baselines on this dataset: 1) the work of \cite{stafylakis2017combining}, 2) the multi-grained spatial-temporal model in \cite{wang2019multi}, 3) the GLMIM method in \cite{zhao2020mutual} and 4) the multi-scale TCN \cite{martinez2020lipreading}.

\textbf{Implementations}\quad
We implement our method in the PyTorch framework \cite{NEURIPS2019_9015}. Experiments are conducted on a server with eight 1080Ti GPUs. It takes around four days to train a single model end-to-end on LRW using one GPU and five days for LRW-1000. Note that this training time is significantly lower than other works \cite{petridis2018end} which requires approximately three weeks per GPU for a training cycle.

\subsection{Results}

\textbf{DC-TCN architectures}\quad
\label{net_struct_explore}
To find an optimal structure of DC-TCN, we first evaluate the impact of various filter sizes $K$ and dilation rates $D$ on LRW while keeping the value of other hyper-parameters fixed. In particular, 
we fix the growth rate $C_o$ and the reduce layer channels $C_r$ to be 64 and 512, respectively, and stack a total of 4 DC-TCN blocks without SE attention. As shown in Table \ref{tab: kd_exploration}, both Fully Dense (FD) and Partially Dense (PD) blocks achieve optimal performance when $K$ and $D$ are set to be $\{3,5,7\}$ and $\{1,2,5\}$, respectively. Therefore, we decide to use this setting for $K$ and $D$ in subsequent experiments. 

\begin{table}[t!]
\small
\centering
\begin{tabularx}{0.9\columnwidth}{s j k k}
\toprule
Filter Sizes $K$ & Dilation Rates $D$ & Acc. (\%, FD) & Acc. (\%, PD)\\ 
\midrule
\{3,5\} & \{1,2,3\} & 86.68 & 86.84 \\ \midrule
\multirow{3}{*}{\textbf{\{3,5,7\}}}  & \{1,2\} & 86.88 & 87.01 \\ 
& \{1,2,4\} & 87.07 & 87.48 \\  
& \textbf{\{1,2,5\}} & \textbf{87.11} & \textbf{87.50} \\ \midrule
\multirow{3}{*}{\{3,5,7,9\}}  & \{1,2\} & 86.99 & 87.26 \\
& \{1,2,4\} & 86.92 & 87.15 \\ 
& \{1,2,5\} & 86.85 & 87.16  \\  
\bottomrule
\end{tabularx}
\vspace{2mm}
\caption{Performance on the LRW dataset of DC-TCN consisting of different filter sizes $K$ and dilation rates $D$. The top-1 accuracy of fully dense (FD) and partially dense (PD) blocks is reported. Other network parameters are fixed, and for simplicity all SE attention is temporarily disabled.}
\label{tab: kd_exploration} 
\end{table}

\begin{table}[t!]
\small
\centering
\begin{tabularx}{0.9\columnwidth}{s s j j }
\toprule
Growth rate $C_o$ & Adding SE & Acc. (\%, FD) & Acc. (\%, PD) \\ \midrule 
64 & - & 87.11 & 87.50  \\ 
64 & \checkmark & 87.40 & 87.91  \\ 
128 & - & 87.64  & 88.13 \\ 
\textbf{128} & \textbf{\checkmark} & \textbf{88.01} & \textbf{88.36}  \\ \bottomrule
\end{tabularx}
\vspace{2mm}
\caption{Performance on the LRW dataset of DC-TCN with different growth rate $C_o$ and using SE or not.
The top-1 accuracy of fully dense (FD) and partially dense (PD) blocks are reported. The filter sizes $K$ and dilation rates $D$ are selected as $\{3,5,7\}$ and $\{1,2,5\}$, respectively, while the reduce layer channels $C_r$ and the total block number $B$ are set to 512 and 4.}
\label{tab: grb_se_exploration}
\vspace{-2mm}
\end{table}

Once the optimal values of $K$ and $D$ are found, we have further investigated the effect of different growth rate $C_o$ settings and the addition of SE block, while the reduce layer channels $C_r$ and the total block number $B$ are set to 512 and 4, respectively. As shown in \ref{tab: grb_se_exploration}, it is evident that: 1. the performance of using 128 for $C_o$ exceeds that of using 64, and 2. the effectiveness of adding SE in the block is validated since it consistently leads to higher accuracy when $C_o$ stays the same.

To sum up, we have selected the following hyper-parameters as the final DC-TCN model configuration for both FD and PD: the filter sizes $K$ and dilation rates $D$ in each block are set to $K=\{3,5,7\}$ and $D=\{1,2,5\}$, with the growth rate $C_o=128$, the reduce layer channel $C_r=512$ and the block number $B=4$, where SE attention is added after each input tensor.

\begin{table}[t!]
\small
\centering
\begin{tabularx}{\columnwidth}{s j k k}
\toprule
Methods  &	Front-end & Back-end &  Acc. (\%) \\ \midrule  
LRW \cite{chung2016lip} & VGG-M & - & 61.1 \\ \midrule
WLAS \cite{chung2017lip} & VGG-M & LSTM & 76.2 \\ \midrule
ResNet+BLSTM \cite{stafylakis2017combining} & 3D Conv + ResNet34 & BLSTM & 83.0 \\ \midrule
End-to-End AVR \cite{petridis2018end} & 3D Conv + ResNet34 & BLSTM & 83.4 \\ \midrule
Multi-grained ST \cite{wang2019multi} & ResNet34 + DenseNet3D & Conv-BLSTM & 83.3 \\ \midrule
Two-stream 3D CNN \cite{weng2019learning} & (3D Conv)$\times$2   & BLSTM  & 84.1 \\ \midrule
ResNet18 + BLSTM\cite{stafylakis2018pushing} & 3D Conv + ResNet18 & BLSTM & 84.3 \\ \midrule
GLMIM \cite{zhao2020mutual} & 3D Conv + ResNet18 & BGRU & 84.4 \\ \midrule
Face cutout \cite{zhang2020can} & 3D Conv + ResNet18 & BGRU & 85.0 \\ \midrule
Multi-modality SR \cite{xu2020discriminative} & 3D ResNet50 & TCN & 84.8 \\ \midrule
Multi-scale TCN \cite{martinez2020lipreading} & 3D Conv + ResNet18  &  MS-TCN  & 85.3 \\  \midrule 
\multirow{2}{*}[-1em]{Ours}  & \multirow{2}{*}[-1em]{\shortstack{3D Conv \\ + ResNet18}}   & DC-TCN (PD) & 
\textbf{88.36} \\ 
 &  & DC-TCN (FD)  & \textbf{88.01} \\\bottomrule
\end{tabularx}
\vspace{0mm}
\caption{A comparison of the performance between the baseline methods and ours on the LRW dataset. We report the best results from the fully dense (FD) and the partially dense (PD) blocks, respectively.}
\label{tab: lrw_results}
\vspace{-2mm}
\end{table}

\begin{table}[t!]
\small
\centering
\begin{tabularx}{\columnwidth}{s j k k}
\toprule
Methods  &	Front-end & Back-end &  Acc. (\%) \\
\midrule
ResNet+LSTM \cite{yang2019lrw} & 3D Conv + ResNet34 & LSTM & 38.2 \\  \midrule
Multi-grained ST \cite{wang2019multi} & ResNet34 + DenseNet3D & Conv-BLSTM & 36.9 \\  \midrule
GLMIM \cite{zhao2020mutual} & 3D Conv + ResNet18 & BGRU & 38.79 \\  \midrule
Multi-scale TCN \cite{martinez2020lipreading} & 3D Conv + ResNet18 & MS-TCN & 41.4 \\ \midrule
\multirow{2}{*}[-1em]{Ours}  & \multirow{2}{*}[-1em]{\shortstack{3D Conv\\ + ResNet18}}   & DC-TCN (PD) & \textbf{43.65} \\ 
 &  & DC-TCN (FD)  & \textbf{43.11} \\ \bottomrule
\end{tabularx}
\vspace{0mm}
\caption{A comparison of the performance between the baseline methods and ours on the LRW-1000 dataset.}
\label{tab: lrw1000_results}
\vspace{2mm}
\end{table}

\begin{table}[t!]
\small
\centering
\begin{tabularx}{\columnwidth}{t i i i i i i}
\toprule
Drop $N$ Frames $\rightarrow$ & $N$=0 & $N$=1 & $N$=2 & $N$=3 & $N$=4 & $N$=5 \\
\midrule 
End-to-End AVR \cite{petridis2018end} \footnotemark & 84.6 & 80.2 & 71.3 & 59.5 & 45.9 & 32.9\\
MS-TCN \cite{martinez2020lipreading} & 85.3 & 83.5 & 81.2 & 78.7 & 75.7 & 71.5 \\
Ours (PD) & 88.4 & 86.2 & 84.0 & 81.0 & 77.5 &73.3 \\
Ours (FD) & 88.0 & 86.4 & 83.6 & 81.3 & 77.7 &73.8 \\
\bottomrule
\end{tabularx}
\vspace{0mm}
\caption{The top-1 accuracy of different methods on LRW where $N$ frames are randomly removed from each testing sequence. }
\label{tab:temporal_robustness} 
\vspace{-2mm}
\end{table}

\textbf{Performance on the LRW and LRW-1000 datasets}\quad
In Table \ref{tab: lrw_results} and \ref{tab: lrw1000_results} we report the performance of our method and various baseline approaches on the LRW and LRW-1000 datasets, respectively.
On LRW, our method has achieved an accuracy of 88.01\,\% (FD blocks) and 88.36\,\% (PD blocks), which is the new state-of-the-art performance on this dataset with an absolution improvement of 3.1\,\% over the current state-of-the-art method \cite{martinez2020lipreading} on the LRW dataset. Besides, our method also produces higher top-1 accuracy (43.65\,\% and 43.11\,\% by using PD and FD, respectively) than the best baseline method \cite{martinez2020lipreading} (41.4\,\%) on LRW-1000, which has further validated the generality and effectiveness of the proposed DC-TCN model.

\footnotetext{3D Conv+ResNet18 as front-end and BGRU as back-end.}

\subsection{Discussion}
\begin{figure}[t!]
  \centering
  \includegraphics[width=0.99\linewidth]{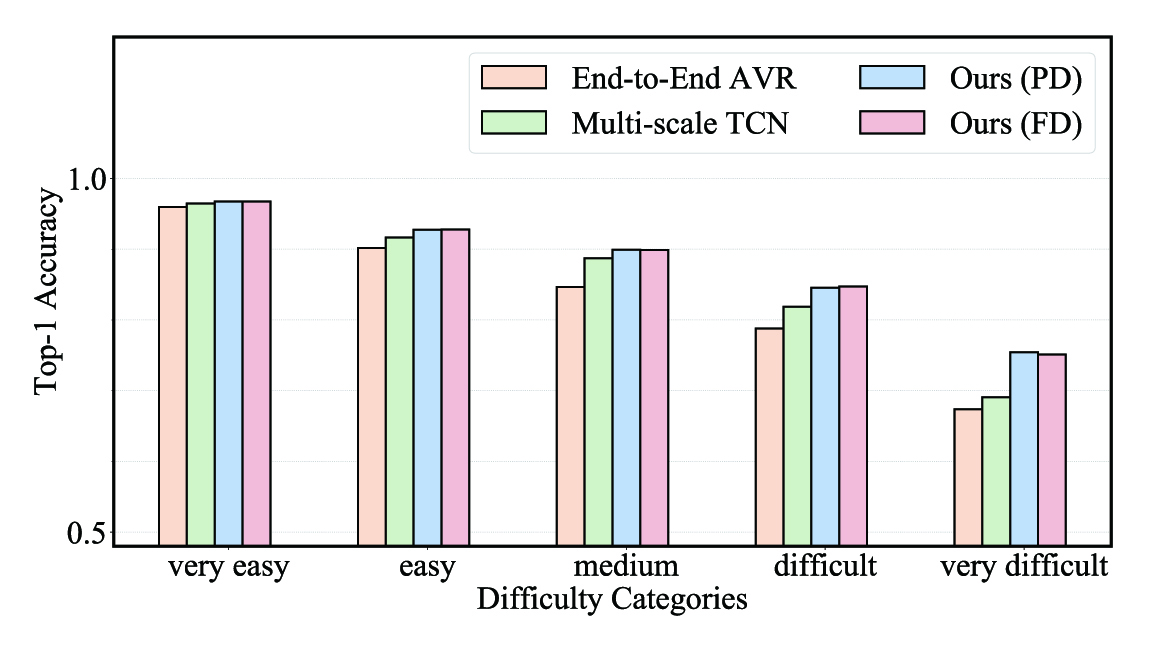}
  \caption{A comparison of our method and two baseline methods (End-to-End AVR \cite{petridis2018end} and Multi-scale TCN \cite{martinez2020lipreading}) on the five difficulty categories of the LRW test set. Our method shows significant improvement over the baselines on these more challenging word classes, which demonstrates that our DC-TCN models can provide more robust temporal features.}
  \label{difficulty_comparsion}
  \vspace{-2mm}
\end{figure}

\textbf{Difficulty Categories}\quad
To intuitively illustrate why our DC-TCN can out-perform the baseline methods, we have examined the classification rates of different methods on five word categories with various difficulty levels. To be specific, we have divided the 500 classes in the LRW test set into five categories (100 words per category) based on their classification difficulty in \cite{martinez2020lipreading}, which are ``very easy'', ``easy'', ``medium'', ``difficult'' and ``very difficult''. Then we compare the performance of our DC-TCN (FD and PD) with two baseline methods (End-to-End AVR \cite{petridis2018end} and Multi-scale TCN \cite{martinez2020lipreading}) on those five difficulty categories, as demonstrated in Fig. \ref{difficulty_comparsion}. We observe that our methods result in slightly better performance than the baselines on the ``very easy'' and ``easy'' categories, however, improvements over the baselines are more significant on the other three groups, especially on the ``difficult'' and the ``very difficult'' categories. Since the improvement of our methods is mainly achieved on those more difficult words, it is reasonably to deduce that our DC-TCN can extract more robust temporal features.

\textbf{Variable Lengths}\quad 
We further evaluate the temporal robustness of different models against video sequences with variable lengths, i.e. $N$ frames are randomly dropped from each testing sequence in LRW dataset where $N$ ranges from 0 to 5. As shown in Table \ref{tab:temporal_robustness}, the performance of End-to-End AVR \cite{petridis2018end} drops significantly as increasing frames are randomly removed from the testing sequences. In contrast, MS-TCN \cite{martinez2020lipreading} and our DC-TCN (both PD and FD) demonstrate better tolerance to such frame removals, mainly due to the usage of variable length augmentation \cite{martinez2020lipreading} during training. Besides, the accuracy of our models (both PD and FD) constantly outperforms that of MS-TCN \cite{martinez2020lipreading} no matter how the number of frames to remove varies, which verifies the superior temporal robustness of our method.  


\section{Conclusion}
We have introduced a Densely Connected Temporal Convolution Network (DC-TCN) for word-level lip-reading in this paper. Characterised by the dense connections and the SE attention mechanism, the proposed DC-TCN can capture more robust features at denser temporal scales and therefore improve the performance of the original TCN architectures. DC-TCN have surpassed the performance of all baseline methods on both the LRW dataset and the LRW-1000 dataset. To the best of our knowledge, this is the first attempt to adopt a densely connected TCN with SE attention for lip-reading of isolated words, resulting in new state-of-the-art performance.
\section*{Acknowledgements}
The work of Pingchuan Ma has been partially supported by Honda and the ``AWS Cloud Credits for Research'' program. The work of Yujiang Wang has been partially supported by China Scholarship Council (No. 201708060212) and the EPSRC project EP/N007743/1 (FACER2VM).

{\small
\bibliographystyle{ieee_fullname}
\bibliography{ref}
}

\end{document}